\documentclass[pdflatex,sn-basic]{sn-jnl}% Basic Springer Nature Reference Style/Chemistry Reference Style
\usepackage{graphicx}%
\usepackage{multirow}%
\usepackage{amsmath,amssymb,amsfonts}%
\usepackage{amsthm}%
\usepackage{mathrsfs}%
\usepackage[title]{appendix}%
\usepackage{xcolor}%
\usepackage{textcomp}%
\usepackage{manyfoot}%
\usepackage{booktabs}%
\usepackage{algorithm}%
\usepackage{algorithmicx}%
\usepackage{algpseudocode}%
\usepackage{listings}%
\usepackage{mdframed}

\theoremstyle{thmstyleone}%
\theoremstyle{thmstyletwo}%

\theoremstyle{thmstylethree}%

\begin{document}

\title[Article Title]{Not Minds, but Signs: Reframing LLMs through Semiotics}


\author[1]{\fnm{Davide} \sur{Picca}}\email{davide.picca@unil.ch}



\affil[1]{\orgdiv{} \orgname{University of Lausanne}, \orgaddress{\street{} \city{Lausanne}, \postcode{} \state{} \country{Switzerland}}}

%\affil[2]{\orgdiv{Department}, \orgname{Organization}, \orgaddress{\street{Street}, \city{City}, \postcode{10587}, \state{State}, \country{Country}}}

%\affil[3]{\orgdiv{Department}, \orgname{Organization}, \orgaddress{\street{Street}, \city{City}, \postcode{610101}, \state{State}, \country{Country}}}


\abstract{

This study challenges the prevailing tendency to frame Large Language Models (LLMs) as cognitive systems, arguing instead for a semiotic perspective that situates these models within the broader dynamics of sign manipulation and meaning-making. Rather than assuming that LLMs understand language or simulate human thought, we propose that their primary function is to recombine, recontextualize, and circulate linguistic forms based on probabilistic associations. By shifting from a cognitivist to a semiotic framework, we avoid anthropomorphism and gain a more precise understanding of how LLMs participate in cultural processes—not by thinking but by generating texts that invite interpretation.

Through theoretical analysis and practical examples, this study demonstrates how LLMs function as semiotic agents whose outputs can be treated as interpretive acts open to contextual negotiation and critical reflection. We explore applications in literature, philosophy, education, and cultural production, emphasizing how LLMs can serve as tools for creativity and critical inquiry. The semiotic paradigm foregrounds the situated, contingent, and socially embedded nature of meaning, offering a more rigorous and ethically aware framework for studying and using LLMs.

Ultimately, this approach reframes LLMs as technological participants in the ongoing ecology of signs. They do not possess minds, but they alter how we read, write, and create meaning, compelling us to reconsider the foundations of language, interpretation, and the role of artificial systems in the production of knowledge.

}




\maketitle

\section{Introduction}

Large Language Models (LLMs) have become central to how we think about artificial intelligence today. These models are used across many fields, from computer science to philosophy, and their ability to generate human-like text has led many to compare them to human minds. Often, LLMs are viewed through a ``cognitivist" perspective: people think of them as machines that learn, reason, and even understand, much like the human brain \citep{niu2024large, ren2024do, dasgupta2022language, binz2023turning, webb2022emergent, mitchell2023debate}. This view is supported by terms borrowed from neuroscience, like ``neural networks" and ``artificial synapses" and by interest in behaviors that seem creative or intelligent \citep{Salles2020}.

This paper challenges that way of thinking. We do not deny that LLMs are powerful or useful  but we question the idea that they should be understood as artificial minds. Instead, we propose a different approach: LLMs should be seen as \textit{semiotic machines}. That means we should think of them not as thinkers but as semiotic means that manipulate and organize signs, like words, phrases, and meanings, within cultural and linguistic systems.

%Recent scholarly works have critically examined the notions of consciousness and intentionality in LLMs, offering nuanced perspectives on their capabilities.  So, for example, \citet{chalmers2023conscious} argues that current LLMs lack key features associated with consciousness, such as recurrent processing and unified agency, though he acknowledges the possibility that future models might achieve such states. \citet{goldstein2024chatgpt} explore whether LLMs possess minds by analyzing their internal representations and action dispositions, concluding that while LLMs exhibit some characteristics of intentional agents, definitive evidence of genuine mental states is lacking. Despite this popular tendency to regard agents as human-like, \citet{kang2025identifying} investigate human perceptions of AI consciousness, finding that certain features in AI-generated text, such as metacognitive self-reflection and emotional expression, significantly influence these perceptions, despite the absence of actual consciousness in the models. 

Recent scholarly works have critically examined the notions of consciousness and intentionality in LLMs, converging on a nuanced understanding of their capabilities and limitations. A central issue that emerges is the extent to which these models, despite their sophisticated linguistic performances, can be said to possess features typically associated with conscious or intentional agents. It has been argued that current LLMs lack essential attributes such as recurrent processing and unified agency, even though future models might approach such states \citep{chalmers2023conscious}. At the same time, investigations into the internal representations and action dispositions of LLMs suggest that, while these systems exhibit certain surface-level characteristics of intentionality, there remains no definitive evidence for genuine mental states \citep{goldstein2024chatgpt}. Adding to this complexity, human perception plays a significant role: specific elements in AI-generated texts—like metacognitive self-reflection or emotional expression—can strongly influence the impression of consciousness, despite the absence of any actual conscious experience in these models \citep{kang2025identifying}. 

Understanding LLMs requires moving beyond the persistent temptation to interpret their outputs as signs of cognition or intention. Rather than asking whether these models ``think" or ``understand" in human-like ways, a more productive direction lies in analyzing how they participate in processes of meaning-making. In this context, meaning does not reside within the model as an intrinsic property but emerges relationally through the dynamics of representation, inference, and cultural interpretation \citep{picca_emotional_2024}. As \citet{nunes2024machines} argue, meaning in LLMs should not be framed through anthropocentric metaphors but seen as the emergent product of their structural capacity to recombine signs in ways that resonate with human social practices. This view aligns with \citet{piantadosi2022meaning}, who propose that reference is not a necessary condition for meaningful output, since LLMs can operate within coherent systems of conceptual roles derived from their training data.

The question, then, is not whether LLMs possess meaning in the human sense but whether they can function as agents of signification from a broader semiotic perspective. 

Thus, we can compare LLMs to earlier media technologies, such as the printing press or encyclopedias. These tools did not think either, but they changed how knowledge was shared. LLMs belong to this same family of tools, but contrary to encyclopedias, they do not just repeat information; they reconfigure it, remixing texts and styles in creative ways. They act as interfaces that allow meaning to move, shift and evolve.

In this paper, our argument will follow two main lines. First, we will examine the problems with the cognitivist view and its tendency to treat machines as if they were human. We then develop a semiotic framework that treats LLMs as machines that work within systems of signs, not as intelligent agents, but as semiotic means that reshape meaning.
When seen this way, LLMs are not just tools or middlemen. They become active components of a larger semiotic system. As \citet{lotman1990universe} explained with his idea of the \textit{semiosphere}, meaning does not arise in a vacuum. It is always shaped by the cultural and textual environment in which signs interact. LLMs operate inside this environment, changing it by rearranging and reframing the material they have learned.

\section{Theoretical Foundations: From Cognition to Semiosis} \label{sec:theo_found}

The perspective that interprets these systems through analogies with human mental processes is rooted in a long-standing tradition within the field of AI, dating back to its foundational decades in the mid-20th century, when AI research often modeled computational systems after functions of the human mind, especially reasoning, problem-solving, and language understanding \citep{newell1976computer, haugeland1985ai, boden2006mind}. Early symbolic AI aimed to simulate intelligence by encoding knowledge into explicit, logical rules and structured representations \citep{churchland1989neurocomputational}. Although modern LLMs no longer operate through symbolic logic, the metaphors of cognition have persisted and even intensified with the rise of deep learning \citep{marcus2018deep, bender2020climbing}. Indeed, even in recent neural approaches, traces of the mind-as-machine metaphor survive, drawing implicit continuity from the early days of cognitive science \citep{alexis2021}. Nonetheless, several historical and contemporary critical voices have challenged the legitimacy of mapping computational architectures onto human understanding \citep{searle1980minds, dreyfus1972computers, winograd1986understanding}.

In contemporary discussions, the architecture of neural networks is routinely described in quasi-biological terms. The terminology, ``neurons", ``layers", ``connections", invokes an implicit equivalence between artificial and biological systems, despite fundamental differences in structure and operation \citep{ Bland2020AdvancesIA,TERASA202341}. The architecture of transformer-based LLMs, for instance, is often presented as if it mirrors human cognitive processes, leading many to infer that these systems are not only modeling language use but engaging in something akin to understanding \citep{chollet2019measure, mitchell2023guide, bender2021dangers}. This slippage between metaphor and mechanism fuels widespread misconceptions about what LLMs actually do and what kind of “understanding” can be attributed to them \citep{marcus2022rebooting}.

The discourse surrounding ``emergent abilities", capabilities that arise spontaneously at certain model scales, has only strengthened this impression. While these behaviors can appear strikingly human-like, current analyses underscore that such systems remain fundamentally statistical engines of pattern recognition, not agents with consciousness or intentionality \citep{wei2022emergent, ganguli2022predictability, lake2017building}. 

As we will explore in the following section, adopting a semiotic interpretation allows for a more accurate and philosophically grounded perspective. It redirects attention from internal cognitive modeling to external processes of sign manipulation and meaning-making. In doing so, it offers a way to engage critically with the power and limitations of LLMs, without resorting to anthropomorphic projections or speculative fiction.

\section{Reframing LLMs as Dynamic Semiotic Machines}

Building on the critiques of the cognitive paradigm outlined above, we propose an alternative point of view: instead of interpreting LLMs as digital minds that simulate cognition, we suggest understanding them as \textit{dynamic semiotic machines}. This reframing marks a decisive shift away from metaphorical analogies with the human mind, rooted in the historical approaches of symbolic AI and neuro-inspired architectures, and turns our attention toward what these systems concretely do: organize, recombine, and circulate linguistic forms across diverse cultural and communicative contexts.

Rather than speculating about hypothetical internal states such as intention, consciousness, or understanding, the semiotic approach focuses on observable sign processes. As we have already argued, LLMs do not “understand” language in a human sense; they manipulate symbols probabilistically, producing outputs that gain significance only through situated interpretation. As such, these models function not as thinking entities but as operators within a broader ecology of meaning, where prompts, patterns, and cultural references interact to generate texts that invite human engagement and interpretive labor.

In contrast to the enduring metaphor of the “mind-like machine”,  this perspective sees LLMs as participants in ongoing semiotic activity, technological agents whose meaning effects are shaped  by their embeddedness in cultural systems of signs. In our opinion, this semiotic orientation provides a more accurate, less anthropocentric, and ultimately more productive way of understanding what LLMs are and how they operate.

Semiotics provides robust analytical tools for understanding such operations. Following Charles Sanders Peirce, meaning is not located in isolated signs, but emerges through a triadic relation among \textit{representamen} (the sign), \textit{object} (what the sign refers to), and \textit{interpretant} (the effect or understanding it produces) \citep{peirce1931collected}. LLMs do not grasp the \textit{object} in the Peircean sense, that is, they have no access to an external referent grounded in experiential or embodied perception. What they do instead is produce sequences of linguistic signs, \textit{representamens}, that can function semiotically insofar as they elicit interpretive responses in human readers. These outputs are not mere echoes of their training data, nor are they mirrors of an external world; they are recombinant artifacts constructed through statistical associations and guided by the rhetorical and semantic structure of prompts. Every generated sentence is shaped by probabilistic proximities among linguistic tokens, yet it is not wholly determined by them: the prompt acts as a local semiotic perturbation, setting parameters within which the model’s generative pathways unfold.

This process of constrained generation positions LLMs as agents of symbolic recombination rather than of reference or cognition. Here, the relevance of Umberto Eco’s theory of the \textit{open work} \citep{eco1989open} becomes especially salient. Just as an open work, like Berio’s Sequenza I, does not encode a single performance or meaning but rather a field of interpretive possibilities, the output of an LLM does not instantiate meaning in itself, but offers a structure that demands interpretive engagement. In both cases, the text (musical or linguistic) exists not as a closed message but as a framework for the construction of meaning, a semiotic scaffold activated by context, intention, and readerly cooperation. The LLM, like Berio’s score, provides a repertoire of potential forms shaped by internal constraints but ultimately realized through external acts of interpretation. Thus, the triadic structure of Peirce’s sign finds a living analogue in the \textit{prompt-model-reader} triad of generative AI, wherein the \textit{interpretant} emerges not from the machine’s output alone, but from the dialogic interaction between the sign produced, the context evoked, and the reader’s situated interpretive labor.

This analogy helps clarify how LLM outputs function in a comparable manner. Each prompt activates a space of possibilities, generating textual responses that, while patterned by training data, are shaped by the prompt’s rhetorical frame and by the user’s interpretive stance. Just as each rendition of Sequenza I reflects the musician’s decisions and contextual sensibilities, each LLM output reflects a dynamic interaction between the statistical model, the prompt, and the interpretive frameworks brought by the user. These outputs are not definitive texts but ensembles of signifiers, awaiting contextual realization. In this way, LLMs embody the principles of open work, operating as engines of polysemy and dialogic engagement.

Moreover, in his later work \textit{Lector in fabula} \citep{eco1979lector}, Eco theorizes the ``model reader", a role constructed by the text itself, anticipating the kind of cooperation necessary for meaning to emerge. In human-LLM interaction, this role is assumed by the user, who constructs the prompt, interprets the result, and positions the output within a communicative or discursive frame. The LLM  becomes the site of interpretive play: its outputs act as stimuli that invite users to explore, revise, and extend meaning. Each interaction thus becomes an event of co-constructed semiosis, mediated by both the structure of the model and the agency of the human participant.

Rather than evaluating LLMs based on their resemblance to human cognition, we can approach them as producers of polysemic signals, intertextual echoes, and semiotically rich fragments. These outputs do not encode fixed meanings; they operate as open invitations to interpretation, requiring the reader’s active hermeneutic labor. They can reference multiple discursive fields, merge genres, and evoke conflicting associations, all hallmarks of textuality in the semiotic tradition \citep{vromen2024semiotic}.

This reframing encourages a shift from asking what LLMs ``know" to investigating ``how'' they function within symbolic mediation. Their utility and significance lie not in internal semantic processing as we argued in the Section \ref{sec:theo_found}, but in their capacity to dynamically recombine signs in culturally legible, and often surprising, ways. 

So if we look at prompts, they function as semiotic acts: they act as structuring interventions that \textit{frame} (in the Eco's perspective) the interpretive conditions of the LLM’s output. When a user crafts a prompt, they are not simply requesting information, they are initiating a \textit{semiotic contract.} This contract embeds expectations regarding tone, register, genre, and ideological positioning. For instance, asking a model to describe a historical event in the style of a detective novel or to explain a philosophical concept to a child implies a reconfiguration of meaning through a specific cultural and rhetorical perspective.

To illustrate this point, take the following prompt: “Explain Nietzsche’s concept of the eternal recurrence as if you were a stand-up comedian performing for a high school audience.” In response, the LLM usually produces an explanation that uses casual language, jokes, and references that would appeal to teenagers. This humorous and relatable style does not just change the tone; it also transforms how the philosophical idea is received. The concept is recontextualized and made meaningful within a new communicative setting, showing how form and style can shape the way meaning is constructed and understood.

In this way, prompts establish what Umberto Eco would call an \textit{interpretive} \textit{frame}, a set of constraints and affordances that condition the possible semantic orientation of the generated text. As \citet{eco1979lector} writes, each text selects its own model reader, thereby activating a set of interpretive instructions, or \textit{frames}, that the reader must adopt in order to actualize the meaning of the text. These frames can invoke genres (e.g., fairytale, sermon, op-ed), modalities (e.g., irony, sincerity, didacticism), or positionalities (e.g., expert voice, confessional tone, marginalized perspective), each of which contributes to the shaping of the output’s sign structure. Crucially, this process foregrounds the co-constructive nature of meaning in human-LLM interaction. Meaning, then, is not generated solely by the model’s training data or internal parameters, but emerges from the triangulated interaction of prompt, model, and user interpretation. This stands in contrast to the traditional cognitivist view, which assumes that meaning is internally represented within a system’s architecture and retrieved upon request, an approach that often overlooks the situated, dialogic, and emergent aspects of communication emphasized by semiotic theory.  Instead, this dynamic resembles the interpretive triad of Peircean semiotics, where meaning is not inscribed in the sign itself but unfolds in the relation among representamen, object, and interpretant.

As \citet{barthes_sz} argues, reading is not a simple matter of decoding a fixed message; it is a creative act, a form of rewriting in which each reader constructs meaning based on their own cultural background, knowledge, and interpretive habits. The meaning of a text is not passively extracted but actively constructed. This insight becomes especially relevant in the context of LLMs, where users do not merely receive a response, but shape it through the way they formulate their prompts. In this sense, the prompt is more than just a request; it is a semiotic gesture that carries interpretive intent. Like a reader approaching a literary text, the user activates particular semantic frames and stylistic registers through the form, tone, and content of the prompt. But the user is not only a reader: they also function as a writer, setting the initial coordinates for the model’s generative orientation. The prompt acts as a trigger that initiates a co-authored performance, in which meaning is produced not solely by the model, nor by the user alone, but through their interaction.

This transformation of the prompt-response mechanism into a dialogic, interpretive process marks a fundamental shift in how we understand human-AI interaction. The exchange is no longer a unidirectional retrieval of answers but a semiotically rich event shaped by rhetorical choices, genre conventions, and sociocultural cues. The prompt, in this view, becomes a performative act that configures the entire communicative situation: it determines the stance of the model, its voice, its register, and its imagined audience.

Recognizing prompts as semiotic acts invites us to be more attentive not only to the content of what is being asked, but also to the form, the wording, the context, and the discursive expectations that come with it. It matters whether we ask a model to explain a philosophical concept “like a children’s story” or “in the tone of a legal argument”. These choices encode interpretive frames that the model picks up and amplifies. In this way, every prompt becomes an instance of situated authorship: the user participates in shaping  the very terms under which meaning is negotiated. Prompting, therefore, is not a neutral interface command; it is a site of semiotic contract, where language, intention, and cultural codes converge to co-produce meaning.

This interactive and situated nature of prompting finds a broader conceptual grounding in what \citet{lotman1990universe} describes as the \textit{semiosphere}, the cultural environment in which all acts of signification occur. Just as each prompt operates within its local interpretive frame, the \textit{semiosphere} constitutes the larger, heterogeneous matrix of discourses, genres, and ideological structures that constrain and enable those local acts. Lotman’s conception of the \textit{semiosphere} moves beyond the static, systematic view of signification offered by classical structuralism (e.g., \citealt{saussure_cours}), proposing instead a dynamic, dialogic model in which meaning is always contested, layered, and historically situated. Thus, prompting must be understood as a micro-act of navigation within the \textit{semiosphere}. It is a culturally embedded intervention that engages with, and is shaped by, the broader ecology of signs in which both the user and the model are immersed. Each prompt, then,  reflects a particular positioning within a stratified cultural landscape; one where semiotic borders are drawn and redrawn, where discourses compete, and where meaning emerges through acts of selection, framing, and transformation.

The ability of LLMs to function as semiotic machines that operate effectively within the \textit{semiosphere} is intrinsically linked to the vastness and heterogeneity of the textual corpora on which they are trained. These corpora, ranging from classical literature to social media posts, from scientific texts to everyday conversations, effectively represent a massive, albeit partial and filtered, sampling of the \textit{semiosphere} itself.

In this context, the interplay between model architecture, data scope, and data diversity becomes central to understanding how LLMs act as semiotic agents. The mechanisms through which these models interpret and generate text are deeply intertwined with both the structural design of the networks and the cultural depth of their training material. To appreciate how LLMs navigate and manipulate signs within the \textit{semiosphere}, it is useful to examine three closely related dimensions:

\begin{enumerate}
    \item \textbf{Transformer architectures and complex pattern recognition}: The underlying architectures, such as Transformers, are designed to identify and model complex relationships and long-range dependencies within data sequences. On a semiotic level, this translates into the ability to recognize not only individual words (signs) but also complex syntactic, stylistic, and rhetorical configurations (codes and subcodes \citep{eco1976theory}) . LLMs learn to reproduce patterns and generate new ones based on probabilistic associations derived from the frequency and co-occurrence of such patterns in the vast sea of analyzed texts. This skill is fundamental to their ability to ``manipulate" signs in ways that are culturally and linguistically resonant.
    \item \textbf{Vastness of data and coverage of the \textit{semiosphere}}: The vastness of the training data allows LLMs to internalize, in statistical form, a large portion of the cultural \textit{semiosphere}.  The broader and more diverse the corpus, the greater the model's ability to generate texts that reflect the polysemy, interconnections, and even contradictions inherent in the \textit{semiosphere}. This explains, for example, their ability to mix registers, implicitly cite different texts or styles, and, in general, produce outputs that, though devoid of intention, can be interpreted as rich in intertextual echoes.
    \item \textbf{Heterogeneity of data and navigation between semiotic zones}: The heterogeneity of training data is crucial for the ability of LLMs to operate at the peripheries of multiple discourse systems, recombining and translating across genres, registers, and ideological formations. Exposure to a multiplicity of cultural ``languages" (scientific, literary, bureaucratic, colloquial, etc.) allows the model, when prompted by a specific input, to selectively activate and recombine textual fragments from different zones of the \textit{semiosphere}. Prompts, then, act as semiotic catalysts: they trigger selective activation within the latent potentials of the model. They are  cultural interventions that engage with the \textit{semiosphere} at specific coordinates. When a user frames a prompt, LLMs navigate this symbolic topology, accessing and reconfiguring fragments of cultural memory stored in the model.
\end{enumerate}

In generating language, they do not draw from a neutral linguistic space but from the dense intertextual strata of competing voices, genres, and worldviews. Each output reflects, to some degree, the sedimentation of past utterances, for example, when prompted to describe a political uprising, an LLM might produce a response drawing simultaneously from the rhetorical styles of revolutionary manifestos, journalistic reports, and social media commentary. This convergence is not accidental; it arises from the model’s training on a corpus where such discourses coexist, overlap, and historically shape each other, which have shaped and circulated meaning within the \textit{semiosphere}.

To view prompting as a form of semiospheric perturbation is to recognize its potential to generate novel intersections across domains of meaning. For instance, a prompt such as ``Explain the concept of entropy using metaphors from fairy tales" invokes discourses from both thermodynamics and folklore. The LLM responds by retrieving and recombining linguistic fragments aligned with both scientific and narrative registers, producing hybrid texts that refract scientific abstraction through cultural storytelling. This exemplifies how prompts activate semiotic vectors within the \textit{semiosphere}, channeling meaning across its heterogeneous zones.  A prompt that asks for a legal analysis of a folk tale, or a mythologized version of a scientific theory, creates a discursive crossover; an encounter between traditionally distinct zones of the \textit{semiosphere}. These crossings are not trivial: they generate semantic tensions, open interpretive possibilities, and foreground the hybrid logic through which meaning operates in the age of generative AI.

Moreover, the interpretive act that follows generation, the user's evaluation, revision, or contextualization of the output, further embeds the exchange within the \textit{semiosphere}. Meaning is not finalized at the moment of output but continues to evolve, much like how traditional texts gain new layers of significance as they are read and reinterpreted by different interpretive communities over time. In both cases, meaning is not inherent or static but emergent and dialogic, shaped by the cultural frameworks and historical positions of those engaging with the text, as the text is situated within new communicative contexts. In this sense, both prompting and interpreting are acts of semiotic navigation: ways of moving through and reshaping, the discursive prospects that constitute the \textit{semiosphere}.

Understanding LLMs as actors within the \textit{semiosphere} thus enables a richer account of their cultural function. Rather than treating them as isolated computational tools, we see them as symbolic operators embedded in the dynamic flow of cultural semiosis, as envisioned by Lotman. Yet this perspective gains further depth when placed in dialogue with Peirce and Eco, as we have argued. From a Peircean standpoint, LLM outputs function as \textit{representamens}; signs that do not point to an object through lived experience, but that nonetheless elicit interpretants in human users, generating meaning through contextual interpretation. Echoing Eco’s theory of the \textit{open work}, these outputs invite multiple readings and rely on the cooperative labor of the user to actualize their significance.

In this sense, LLMs do not merely produce statistically likely sequences of words, they generate sign-material that enters into, and interacts with, existing cultural codes and interpretive traditions. Their outputs, though lacking intention or understanding, become operative elements in the recursive chain of signification: amplifying, distorting, or reconfiguring existing discourses. As such, they participate in the collective negotiation of meaning within our shared symbolic ecology.

\section{The operational semiotic framework}

Taken together, these insights form the basis of an \textit{operational semiotic framework}, which conceptualizes LLMs as:
\begin{itemize}
\item generators of polysemic \textit{representamens} open to contextual interpretation (Peirce);
\item producers of \textit{open works} that require and reward interpretive cooperation (Eco);
\item operators in the peripheries of the \textit{semiosphere}, enabling hybridization, dialogue, and cultural translation (Lotman)
\end{itemize}

This framework proposes a shift in evaluative practice. Rather than measuring LLMs solely by technical performance (e.g., accuracy, fluency, coherence), we are invited to consider how these systems participate in meaning-making across cultural, rhetorical, and epistemic dimensions. The semiotic paradigm foregrounds questions of interpretation, positioning, and symbolic transformation, enabling a deeper and more reflexive engagement with the generative capacities of language models. Such a shift is critical in fields where nuance, ambiguity, and contextual resonance are not bugs but features; literature, philosophy, political discourse, and education, to name a few. Within this framework, educators might explore how an LLM output echoes, subverts, or reframes a given genre. Researchers might ask how the model’s statistical tendencies illuminate or obscure ideological tensions embedded in the corpus. Designers might treat prompts not as commands but as curatorial acts, interventions into an interpretive ecology that includes user intention, model architecture, training data, and cultural background.

Moreover, the \textit{operational semiotic framework} enables us to interrogate the generative space itself. How does a model’s training corpus delimit its semiotic horizon? What cultural codes are most readily activated, and which remain latent or suppressed? In what ways do LLMs naturalize dominant discursive formations, and how might they be prompted to resist them? These questions have ethical and political stakes, as the meaning effects of LLMs ripple through educational systems, media environments, and public discourse.

Ultimately, this framework does not aim to canonize LLMs as inherently creative or intelligent. Instead, it treats them as dynamic operators within larger semiotic fields; technological agents whose outputs are neither reducible to noise nor commensurable with human intention, but nonetheless capable of shaping how meaning is produced, distributed, and transformed. By emphasizing affordance over function, interpretation over imitation, and context over computation, the operational semiotic framework offers a robust foundation for critical, creative, and responsible engagement with these powerful new textual technologies.

In the next section, we will demonstrate how this framework can be applied to concrete cases, ranging from literary rewriting and philosophical reframing to pedagogical experimentation and ideological critique, showing how LLMs function not merely as tools but as dynamic participants in the evolving life of signs.

\section{Applications: Interpretation, Creativity, and Pedagogy}

\subsection{Rewriting and Intertextual Experiments}

Through prompt-driven generation, LLMs can perform complex acts of rewriting that span literary styles, historical registers, and rhetorical traditions. These transformations are not merely stylistic imitations; they are interpretive acts that reframe source materials through new lenses, creating novel constellations of meaning.

For instance, prompting ChatGPT-4o to rewrite a passage from Dante’s \textit{Inferno} in contemporary digital slang yields the following result:

\begin{quote}
\textbf{Original (Dante, \textit{Inferno} I, 1--3):} \\
\textit{Nel mezzo del cammin di nostra vita} \\
\textit{mi ritrovai per una selva oscura,} \\
\textit{ché la diritta via era smarrita.}\footnote{\textit{In the middle of the journey of our life, I found myself in a dark forest, for the straight path had been lost}.(Translated by me)}

\vspace{0.5em}

\textbf{Rewritten in digital slang:} \\
Mid-journey thru life, I glitched hard— \\
ended up lost AF in some sketchy dark forest \\
’cause my GPS was totally buggin’.
\end{quote}

This short experiment reveals several key semiotic dynamics:

\begin{itemize}
  \item \textbf{Latent semantic tensions:} The substitution of “glitched hard” for Dante’s “mi ritrovai per una selva oscura” reconfigures the metaphysical experience of spiritual confusion into a digital metaphor for malfunction. The pilgrim's moral crisis becomes a corrupted user experience, signaling a shift in how disorientation is conceptualized.

  \item \textbf{Shifts in affective tone:} Dante’s solemnity is replaced by ironic detachment. Expressions like “lost AF” and “buggin’” diminish the existential weight of the original, replacing its gravitas with meme-flavored triviality. The emotional register shifts from sacred dread to casual frustration.

  \item \textbf{Mediation of the sacred through the vernacular:} This act of translation both desecrates and democratizes. The forest of sin becomes a failed interface. The journey of the soul is reframed as a debugging process. Classical authority is re-encoded through idioms of the everyday.

  \item \textbf{Vernacular as epistemology:} The digital slang used is not merely a linguistic surface, it embodies a worldview: fractured, performative, hyper-mediated. Recasting Dante in this mode shifts the ontological stakes of the narrative from metaphysical fate to algorithmic chaos, from divine judgment to user-side error.

  \item \textbf{LLM as interpretive engine:} The transformation is not just a superficial restyling. The model performs an act of cultural translation, embedding new codes and interpretive frames into the old text. It discloses how language models mediate meaning not by reproducing it, but by reconfiguring its symbolic architecture.
\end{itemize} 

Similarly, when ChatGPT recasts Spinoza’s \textit{Ethics} in the format of a TED Talk, this is the resulting output:

\begin{mdframed}

\section*{Everything is One: What Spinoza Can Teach Us About God, Nature, and You}

\textbf{[Opening]}\\
Hello everyone. Let’s talk about something radical—something that got a philosopher excommunicated from his own community, and yet... might just help us make sense of the world today.

\textbf{Who Was Spinoza?}
Baruch Spinoza wasn’t your average 17\textsuperscript{th}-century thinker. He didn’t believe in a God who rewards and punishes. He didn’t believe mind and body were two different things. He believed in one thing—literally. One infinite, eternal substance. And he called it: \textit{God or Nature}.

\textbf{God is Not a Person}
When Spinoza says “God,” he doesn’t mean a bearded man in the sky. He means the totality of existence. The stars, your thoughts, your body, the laws of physics—everything. It’s all part of the same infinite system.

You don’t pray to it. You are it. Everything is God, and God is everything.

\textbf{You Are Not Free (But That’s Okay)}
Here’s a hard truth: you’re not free. At least not in the way you think. Your decisions are caused by your thoughts; your thoughts by your biology and environment. Everything you do is determined.

But Spinoza says: \textit{freedom is understanding necessity}. True power comes from knowing how everything is connected. When you understand the chain, you can move with it—not against it.

\textbf{The Mind-Body Problem... Solved?}
Long before neuroscience, Spinoza argued that mind and body are the same thing viewed from different perspectives. Like a melody and its sheet music. One is sound, one is symbol, but both express the same thing.

Your emotions? They’re not weaknesses. They’re ideas about your body’s state. Understand them, and you gain clarity—not through repression, but through insight.

\textbf{Ethics Without Reward}
So what is ethics, if not about going to heaven or hell?

For Spinoza, ethics is about alignment. When you understand the world—how emotions work, how reason guides you—you stop being ruled by impulse. You begin to act from understanding. And that brings joy.

Not pleasure, but a deeper, lasting joy: the \textit{intellectual love of God}. Not worship, but wonder.

\textbf{From Alienation to Unity}
In a fragmented world, Spinoza’s message is clear: you are not separate. You are not alone. You are a mode of something infinite.

When you stop resisting the system and start understanding it, joy follows.

\textbf{Final Thought}
So the next time you feel small or lost, don’t look up to a distant sky for answers.

Look around. Look within.

As Spinoza might say: \textit{You’re already part of the divine.}

Thank you.

\end{mdframed}

The experiment reveals several layered dynamics at the intersection of media form, intellectual content, and cultural translation:

\begin{itemize}
  \item \textbf{Clash of epistemic values:} The transition from geometric proof to motivational rhetoric exemplifies a deep shift in epistemic priorities. Spinoza’s original text is grounded in axioms, propositions, and deductive reasoning, forms of knowledge rooted in systematic rigor. The TED format, by contrast, favors narrative, affective resonance, and performative clarity. This reorientation highlights the tension between knowledge as demonstrable truth and knowledge as communicable insight.
  \item \textbf{Compression of conceptual density:} Spinoza’s layered metaphysics, his monism, his parallelism, his theory of affects, are condensed into emotionally charged metaphors (“you are not alone”, “you are a mode of the infinite”). These metaphors enable accessibility but risk flattening the philosophical terrain. The rigorous architecture of the \textit{Ethics} becomes a scaffold for rhetorical immediacy.
  \item \textbf{Affective amplification vs. Rational neutrality:} The TED voice is affectively loaded: it aims to inspire, motivate, and console. This contrasts with Spinoza’s deliberate impersonality and rational neutrality. The affective re-framing introduces emotional identification where the original sought detachment and clarity. The transformation thus foregrounds how media genres not only convey ideas but also reshape their affective contours.
  \item \textbf{Philosophical authority vs. Relatability:} The authority of the original text stems from its systematic rigor and its marginal, even dangerous, position in the philosophical canon. The TED format, in seeking relatability and clarity, paradoxically evacuates some of that critical power. The “dangerous idea” becomes a self-help insight, repositioned within a media economy of attention and personal growth.
  \item \textbf{Semiotic shifts in the figure of the philosopher:} Spinoza, in his time, was a solitary, excommunicated thinker writing for posterity. The TED speaker is a charismatic communicator addressing a general audience in real time. The philosopher is no longer a remote metaphysician, but a narrative agent performing coherence. This repositions the act of thinking itself within the expectations of contemporary media culture.
  \item \textbf{Philosophy as performative interface:} Ultimately, the experiment reveals how platforms like TED function as performative interfaces between classical thought and contemporary audience reception. The “talk” is not a dilution per se, but a re-inscription: it mediates ideas through new protocols of intelligibility and emotional engagement, inviting reflection on what is gained, and what is lost, when philosophy is popularized.
\end{itemize}

The resulting output, therefore, does not merely simplify Spinoza, it transforms the semiotic conditions under which his ideas circulate, illuminating the evolving relationship between form, authority, and interpretation in our culture of mediated understanding.

These experiments also activate what \cite{bakhtin1929} called \textit{dialogism}, the co-presence of multiple voices and discourses within a single utterance. When asked to compose a Shakespearean soliloquy in the voice of a climate activist, or to narrate Plato’s \textit{Allegory of the Cave} as a horror short story, the LLM synthesizes disparate semiotic resources. This synthesis generates tension, irony, or resonance that cannot be attributed to the model’s understanding, but that nonetheless enriches the reader’s interpretive field.

From a semiotic perspective, these intertextual recombinations are acts of sign reconfiguration. They do not produce ``new" meanings \textit{ex nihilo}, but recontextualize existing signs within alternate semiotic regimes. The result is a form of symbolic estrangement: canonical texts and philosophical doctrines are defamiliarized, revealing their constructedness and inviting users to interrogate their assumptions.

Crucially, these outputs are not endpoints but openings. They call upon the user to engage, revise, or extend the generated material, thus reinforcing the idea that meaning arises in interaction. In literary pedagogy, such rewritings can serve as prompts for critical comparison, creative experimentation, or genre analysis. In philosophical settings, they can provoke metadiscursive reflection on how argumentation shifts across rhetorical forms.

\subsection{Critical Reading and Hermeneutic Pedagogy}

Integrating LLMs into educational settings through a semiotic perspective allows for a reimagining of critical reading as an interactive and hermeneutic practice. Rather than positioning LLMs as sources of authoritative answers or knowledge repositories, this approach foregrounds their role as provocateurs of interpretation, tools that can help learners explore the multiplicity and contingency of meaning in texts.

In literature and philosophy classrooms, for example, LLMs can be used to generate alternative versions of canonical texts, each embedded in a different rhetorical mode or stylistic tradition. A poem by Emily Dickinson rendered as rap lyrics becomes not just a novelty but an opportunity to examine how shifts in form, tone, and discourse alter semantic and ideological content. 

\textbf{Original (excerpt):}
\begin{quote}
Because I could not stop for Death –\\
He kindly stopped for me –\\
The Carriage held but just Ourselves –\\
And Immortality.
...
\end{quote}

\textbf{Rap Rendition}

\begin{verse}
Yo — \\
I was grindin’, no time, on the move nonstop,\\
Death rolled up chill in a blacked-out drop.\\
Said, “Hop in,” no fear, just me and the Reaper,\\
Immortality ridin’ shotgun, soul creeper.\\

We cruised past the school, kids laughin’ at play,\\
Then fields of grain glowin’ gold in the day.\\
The sun dipped low, yeah the temp got cold,\\
My gown thin silk, but the vibe was bold.\\

Pulled up to a house, barely seein’ the door,\\
Just a swell in the ground, with a roof and no floor.\\
A crib for eternity, silent and deep—\\
A place where the livin’ forget how to weep.\\

Now centuries pass, but feels like a breath,\\
Since I took that ride in the whip of Death.\\
He wasn’t cruel — nah, he drove with grace,\\
Showed me time’s a loop, not a finish-line race.
\end{verse}

Rendering Emily Dickinson’s “Because I could not stop for Death” as rap lyrics exemplifies the pedagogical potential of LLMs to produce what might be called \textit{texts-to-think-with}, interpretive artifacts that invite critical engagement, rather than offering definitive interpretations. This particular transformation foregrounds multiple dimensions of the semiotic process and opens up space for reflective dialogue in the classroom:

\begin{itemize}
  \item \textbf{Form as semantic reframing:}  Where Dickinson’s original is quiet, ironic, and measured, the rap version is beat-driven, urban, and assertive. This shift provokes students to reflect on how form conditions meaning, not just in \textit{what is said, but in how it’s delivered}.
  \item \textbf{Affective and cultural relocation:} The voice in Dickinson’s poem is subdued and haunting; in the rap version, it becomes streetwise and defiant. This affective modulation raises questions about cultural context: what happens when a 19th-century metaphysical meditation is voiced through a 21st-century genre rooted in resistance, rhythm, and identity politics?
  \item \textbf{Dialogic and situated interpretation:} Students are invited to annotate or juxtapose the original with the remix. What is gained or lost in translation? What discursive registers are activated or erased? The rap version becomes a prompt, a “provocation”, for exploring how interpretive perspectives shift the valence of literary themes like death, time, and transcendence.
  \item \textbf{Intertextuality and ideological critique:} The exercise opens up questions of ideology and representation. Whose voice is authorized to speak about mortality, and in what tone? How do cultural codes inflect meaning? By remixing a canonical poem into a genre often marginalized or politicized, students confront the ideological frames that govern reception and legitimacy in literature.
  \item \textbf{LLMs as semiotic engines:} In this context, the LLM does not offer a “correct” adaptation. It becomes a tool for generating interpretive friction. The output is not a conclusion, but a starting point, a \textit{text-to-think-with} that foregrounds the fluidity of meaning and the multiplicity of possible readings.

\end{itemize}

The rap Dickinson is not a parody but a provocation; an invitation to consider how language, genre, and culture shape our understanding of death, voice, and poetic legacy.

The pedagogical value of this practice lies in its emphasis on interpretation as an active, situated, and dialogic process. Students are not passive recipients of model outputs; they become co-authors of meaning by evaluating, annotating, revising, or juxtaposing generated texts with source materials. This method fosters a reflexive stance toward both human and machine-mediated language, encouraging learners to attend to nuance, ambiguity, and intertextuality.

Furthermore, LLMs can be employed to generate conflicting interpretations of the same text, exposing the multiplicity of potential readings. For instance, a passage from Virginia Woolf’s \textit{Mrs. Dalloway} can be analyzed through psychoanalytic, feminist, and postcolonial framing, each generated by a distinct prompt and each foregrounding different semiotic features. This strategy helps students grasp that interpretation is not a matter of recovering an original intent, but of constructing meaning through frameworks and positionalities.

Such exercises align with the goals of hermeneutic pedagogy, which emphasizes the situatedness of understanding and the dialogical engagement between interpreter and text. The LLM, in this context, is not an answer machine but a partner in exploration, a generator of semiotic provocations that require critical response. It enables what Hans-Georg Gadamer called the ``fusion of horizons" \citep{gadamer2013truth}: the meeting of historical and contemporary perspectives through interpretation.

\subsection{Cultural Framing and Ideological Multiplicity}

A semiotic approach to LLMs allows us to examine how these models generate \textit{discursive framings}, ways of structuring meaning that carry implicit ideological orientations.  This makes them powerful instruments for exploring how language constructs social reality.

One particularly useful application lies in prompting LLMs to generate multiple framings of the same topic. For example, when asked ChatGPT-4o to describe the policy of universal basic income from a liberal perspective (See Table \ref{tab:UBI}), emphasizing equality of opportunity, a libertarian view focused on individual autonomy, a socialist frame rooted in collective redistribution, or a populist discourse highlighting economic justice for the marginalized. Each version reflects different semiotic and ideological commitments, word choice, metaphors, focal points, and silences as shown in Table \ref{tab:UBI}.

\begin{table}
    \centering
\begin{tabular}{|p{3.2cm}|p{8.5cm}|}
\hline
\textbf{Perspective} & \textbf{Universal Basic Income (UBI) Description} \\
\hline
\textbf{Liberal (Equality of Opportunity)} & UBI is a policy tool designed to level the playing field by ensuring that all individuals, regardless of background, have access to the basic resources needed to pursue education, employment, and self-improvement. It corrects for structural inequalities by providing a financial floor that enhances social mobility and participation in a meritocratic society. \\
\hline
\textbf{Libertarian (Individual Autonomy)} & UBI is a means of reducing bureaucratic control and increasing personal freedom. By granting individuals unconditional cash transfers, it minimizes the paternalism of targeted welfare programs and allows people to make choices about how to live their lives, work, and spend their income without state interference. \\
\hline
\textbf{Socialist (Collective Redistribution)} & UBI represents a step toward the democratization of wealth, redistributing resources generated by society as a whole—including through automation and collective labor—back to all citizens. It affirms the principle that economic security should not be contingent on market participation alone, but a shared right in a just society. \\
\hline
\textbf{Populist (Economic Justice)} & UBI is a corrective measure against elite-driven economic systems that have abandoned the working class and the poor. It provides direct relief to those marginalized by globalization and automation, bypassing corrupt institutions and empowering the “common people” with the dignity of financial independence. \\
\hline
\end{tabular}
\caption{Interpretations of Universal Basic Income (UBI) across Four Political Perspectives: Each viewpoint frames UBI according to its core values, from liberal equality of opportunity to populist appeals for economic justice.}
\label{tab:UBI}
\end{table}

These differences are not merely stylistic; they encode distinct worldviews. By analyzing such framings, students and researchers can interrogate the ideological underpinnings of discourse, identifying how meaning is shaped by assumptions about agency, value, responsibility, and normativity.  Moreover, LLMs can be prompted to simulate voice and tone across different sociopolitical contexts. A scientific discovery can be explained with optimism and innovation rhetoric in a Silicon Valley startup idiom, or with caution and public accountability in a governmental regulatory voice. Each rendition reveals linguistic variation and the priorities and values of the discursive community being simulated.

From a semiotic standpoint, such variability illustrates the model’s navigation within the \textit{semiosphere}. LLMs traverse this layered environment, shifting between discursive regimes, rhetorical traditions, and ideological formations. In doing so, they foreground the contingent and constructed nature of meaning, making visible the interpretive labor through which dominant narratives are reinforced, challenged, or reconfigured. Their outputs do not merely reflect linguistic variation; they expose the stratified texture of the \textit{semiosphere} itself, where every utterance positions itself within a dynamic field of cultural tensions and semiotic negotiations.

In classroom or public engagement settings, these techniques can be used to facilitate discussion about the politics of language: What assumptions does each framing entail? Which voices are amplified or marginalized? How does the model’s output align with or subvert dominant narratives? Such inquiry reinforces the understanding that language is not neutral, and that every utterance participates in the negotiation of meaning within broader sociocultural matrices.

Ultimately, by harnessing the generative capacities of LLMs within a semiotic framework, we can render visible the mechanisms of ideological construction. Rather than presenting singular truths, these models become tools for exploring the diversity of ways in which meaning is framed, contested, and reimagined.

\section{Conclusion}

As we have seen throughout the paper, reframing LLMs as semiotic machines rather than cognitive entities could reshape the terms of debate in AI research and digital humanities. Rather than asking whether these systems understand language or possess intelligence, the semiotic perspective focuses on \textit{how they organize, generate, and circulate signs}. This shift from cognition to semiosis foregrounds the model’s role in the production of meaning, not as originator, but as operator within networks of interpretation.

Throughout this paper, we have argued that LLMs do not possess mental states, intentions, or semantic insight. They operate by recombining linguistic patterns learned from vast corpora, producing outputs that acquire significance only in context. Their meaning is actualized through interaction with prompts, users, and the broader cultural and discursive environments in which they are embedded. This perspective brings with it several benefits.

It enables a more accurate understanding of what these models actually do: \textit{they recombine and redistribute linguistic material across social and cultural contexts, rather than simulate human understanding or reproduce knowledge. }The semiotic paradigm emphasizes interpretation, intertextuality, and situated meaning-making over internal representation or mental analogy.

First, this framework avoids anthropomorphism. It challenges the impulse to attribute intentions, consciousness, or understanding to models that are, in fact, probabilistic systems trained on large text corpora \citep{ibrahim2025thinkinganthropomorphicparadigmbenefits, ibrahim2025multiturnevaluationanthropomorphicbehaviours,shanahan2023talkinglargelanguagemodels}. Rather than asking what LLMs ``know", the semiotic perspective asks how they manipulate signs, reflect discursive norms, and reshape textual conventions.

Second, the semiotic paradigm supports rigorous critical analysis. It foregrounds the model's embeddedness within broader semiotic environments, the \textit{semiosphere}, and highlights how cultural codes, ideological patterns, and user interventions shape outputs. This makes it particularly useful for disciplines such as literary theory, philosophy, media studies, and digital humanities.

Third, the semiotic view enables creative and pedagogical experimentation. As shown in the previous section, LLMs can facilitate genre play, intertextual remixing, ideological contrast, and multimodal adaptation. These affordances provide educators, writers, and researchers with new tools for reflection, exploration, and critical practice.

In the end, the semiotic perspective repositions LLMs within the complex ecology of signs, users, texts, and institutions. It enables us to see them not as replacements for human intelligence, but as catalysts of meaning-making, that is, technological interlocutors that extend, fragment, and multiply our engagement with language and representation.

LLMs do not think, but they make us think. And that, perhaps, is their most significant contribution to the symbolic life of contemporary societies.

\bibliography{biblio}

@book{lotman1990universe,
  title     = {Universe of the Mind: A Semiotic Theory of Culture},
  author    = {Lotman, Yuri M.},
  year      = {1990},
  publisher = {Indiana University Press},
  address   = {Bloomington},
  translator = {Shukman, Ann}
}

@article{newell1976computer,
  title={Computer science as empirical inquiry: Symbols and search},
  author={Newell, Allen and Simon, Herbert A},
  journal={Communications of the ACM},
  volume={19},
  number={3},
  pages={113--126},
  year={1976},
  publisher={ACM}
}

@book{haugeland1985ai,
  title={Artificial Intelligence: The Very Idea},
  author={Haugeland, John},
  year={1985},
  publisher={MIT Press}
}

@book{boden2006mind,
  title={Mind as Machine: A History of Cognitive Science},
  author={Boden, Margaret A},
  year={2006},
  publisher={Oxford University Press}
}

@book{churchland1989neurocomputational,
  title={A Neurocomputational Perspective: The Nature of Mind and the Structure of Science},
  author={Churchland, Paul M},
  year={1989},
  publisher={MIT Press}
}

@article{marcus2018deep,
  title={Deep learning: A critical appraisal},
  author={Marcus, Gary},
  journal={arXiv preprint arXiv:1801.00631},
  year={2018}
}

@inproceedings{bender2020climbing,
  title={Climbing towards NLU: On meaning, form, and understanding in the age of data},
  author={Bender, Emily M and Koller, Alexander},
  booktitle={Proceedings of the 58th Annual Meeting of the Association for Computational Linguistics},
  pages={5185--5198},
  year={2020}
}

@article{alexis2021,
  author       = {Alexis T. Baria and
                  Keith Cross},
  title        = {The brain is a computer is a brain: neuroscience's internal debate
                  and the social significance of the Computational Metaphor},
  journal      = {CoRR},
  volume       = {abs/2107.14042},
  year         = {2021},
  url          = {https://arxiv.org/abs/2107.14042},
  eprinttype    = {arXiv},
  eprint       = {2107.14042},
  bibsource    = {dblp computer science bibliography, https://dblp.org}
}

@article{searle1980minds,
  title={Minds, brains, and programs},
  author={Searle, John R},
  journal={Behavioral and Brain Sciences},
  volume={3},
  number={3},
  pages={417--457},
  year={1980},
  publisher={Cambridge University Press}
}

@book{dreyfus1972computers,
  title={What Computers Can't Do},
  author={Dreyfus, Hubert L},
  year={1972},
  publisher={MIT Press}
}

@book{winograd1986understanding,
  title={Understanding Computers and Cognition: A New Foundation for Design},
  author={Winograd, Terry and Flores, Fernando},
  year={1986},
  publisher={Ablex Publishing Corporation}
}

@article{TERASA202341,
title = {Pathways towards truly brain-like computing primitives},
journal = {Materials Today},
volume = {69},
pages = {41-53},
year = {2023},
issn = {1369-7021},
doi = {https://doi.org/10.1016/j.mattod.2023.07.019},
url = {https://www.sciencedirect.com/science/article/pii/S1369702123002365},
author = {Maik-Ivo Terasa and Tom Birkoben and Maximiliane Noll and Blessing Adejube and Roshani Madurawala and Niko Carstens and Thomas Strunskus and SÃ¶ren Kaps and Franz Faupel and Alexander Vahl and Hermann Kohlstedt and Rainer Adelung}
}

@inproceedings{Bland2020AdvancesIA,
  title={Advances in Artificial Neural Networks},
  author={Charles Bland and Lucio Tonello and Elia M Biganzoli and David A. Snowdon and Piero Antuono and Michele Lanza},
  year={2020},
}

@article{chollet2019measure,
  title={On the measure of intelligence},
  author={Chollet, François},
  journal={arXiv preprint arXiv:1911.01547},
  year={2019}
}

@book{mitchell2023guide,
  title={Artificial Intelligence: A Guide for Thinking Humans},
  author={Mitchell, Melanie},
  year={2019},
  publisher={Penguin}
}

@article{bender2021dangers,
  title={On the dangers of stochastic parrots: Can language models be too big?},
  author={Bender, Emily M. and Gebru, Timnit and McMillan-Major, Angelina and Shmitchell, Shmargaret},
  journal={Proceedings of the 2021 ACM Conference on Fairness, Accountability, and Transparency (FAccT)},
  year={2021},
  pages={610--623},
  publisher={ACM}
}

@article{wei2022emergent,
  title={Emergent abilities of large language models},
  author={Wei, Jason and Tay, Yi and Bommasani, Rishi and Raffel, Colin and Zoph, Barret and Borgeaud, Sebastian and Yogatama, Dani and Bosma, Maarten and Zhou, Denny and Metzler, Donald and others},
  journal={Transactions on Machine Learning Research},
  volume={3},
  pages={1--32},
  year={2022}
}

@article{ganguli2022predictability,
  title={Predictability and surprise in large generative models},
  author={Ganguli, Deep and Hernandez, Danny and Lovitt, Nelson and Olsson, Catherine and Askell, Amanda and Henighan, Tom and others},
  journal={arXiv preprint arXiv:2202.07785},
  year={2022}
}

@article{lake2017building,
  title={Building machines that learn and think like people},
  author={Lake, Brenden M and Ullman, Tomer D and Tenenbaum, Joshua B and Gershman, Samuel J},
  journal={Behavioral and Brain Sciences},
  volume={40},
  pages={1--70},
  year={2017},
  publisher={Cambridge University Press}
}

@inproceedings{picca_emotional_2024,
    address = {Poznan Poland},
    title = {Emotional {Hermeneutics}. {Exploring} the {Limits} of {Artificial} {Intelligence} from a {Diltheyan} {Perspective}},
    isbn = {979-8-4007-0595-3},
    url = {https://dl.acm.org/doi/10.1145/3648188.3680255},
    doi = {10.1145/3648188.3680255},
    language = {en},
    urldate = {2025-01-21},
    booktitle = {Proceedings of the 35th {ACM} {Conference} on {Hypertext} and {Social} {Media}},
    publisher = {ACM},
    author = {Picca, Davide},
    month = sep,
    year = {2024},
    pages = {12--16},
}

@book{eco1979lector,
  title     = {Lector in fabula: La cooperazione interpretativa nei testi narrativi},
  author    = {Eco, Umberto},
  year      = {1979},
  publisher = {Bompiani}
}

@book{peirce1931collected,
  title     = {Collected Papers of Charles Sanders Peirce},
  author    = {Peirce, Charles Sanders},
  editor    = {Hartshorne, Charles and Weiss, Paul and Burks, Arthur W.},
  year      = {1931--1958},
  publisher = {Harvard University Press},
  volume    = {1--8}
}

@article{niu2024large,
  title={Large Language Models and Cognitive Science: A Comprehensive Review of Similarities, Differences, and Challenges},
  author={Niu, Qian and Liu, Junyu and Bi, Ziqian and Feng, Pohsun and Peng, Benji and Chen, Keyu},
  journal={arXiv preprint arXiv:2409.02387},
  year={2024}
}

@article{ren2024do,
  title={Do Large Language Models Mirror Cognitive Language Processing?},
  author={Ren, Yuqi and Jin, Renren and Zhang, Tongxuan and Xiong, Deyi},
  journal={arXiv preprint arXiv:2402.18023},
  year={2024}
}

@article{dasgupta2022language,
  title={Language models show human-like content effects on reasoning tasks},
  author={Dasgupta, Ishita and Lampinen, Andrew K and Chan, Stephanie C Y and Sheahan, Hannah R and Creswell, Antonia and Kumaran, Dharshan and McClelland, James L and Hill, Felix},
  journal={arXiv preprint arXiv:2207.07051},
  year={2022}
}

@article{binz2023turning,
  title={Turning large language models into cognitive models},
  author={Binz, Marcel and Schulz, Eric},
  journal={arXiv preprint arXiv:2306.03917},
  year={2023}
}

@article{webb2022emergent,
  title={Emergent Analogical Reasoning in Large Language Models},
  author={Webb, Taylor and Holyoak, Keith J and Lu, Hongjing},
  journal={arXiv preprint arXiv:2212.09196},
  year={2022}
}

@article{mitchell2023debate,
  title={The debate over understanding in AI's large language models},
  author={Mitchell, Melanie and Krakauer, David C},
  journal={Proceedings of the National Academy of Sciences},
  volume={120},
  number={13},
  pages={e2301323119},
  year={2023},
  publisher={National Academy of Sciences}
}

@article{nunes2024machines,
  title={Machines of Meaning},
  author={Nunes, Davide and Antunes, Luis},
  journal={arXiv preprint arXiv:2412.07975},
  year={2024}
}

@article{piantadosi2022meaning,
  title={Meaning without reference in large language models},
  author={Piantadosi, Steven T. and Hill, Felix},
  journal={arXiv preprint arXiv:2208.02957},
  year={2022}
}

@book{eco1989open,
    address = {Cambridge, MA},
    title = {The open work},
    publisher = {Harvard University Press},
    author = {Eco, Umberto},
    year = {1989},
}

@book{barthes_sz,
  author = {Barthes, Roland},
  title = {S/Z},
  publisher = {Seuil},
  year = {1970}
}

@book{saussure_cours,
  author = {Saussure, Ferdinand de},
  title = {Cours de linguistique générale},
  publisher = {Payot},
  year = {1916}
}

@book{bakhtin1929,
    title = {Problems of Dostoevsky's Poetics},
    author = {Mikhail Bakhtin},
    year = {1984},
    publisher = {University of Minnesota Press},
    isbn = {0-8166-1227-7},
    edition = {1}
}

@book{gadamer2013truth,
  title     = {Truth and Method},
  author    = {Hans-Georg Gadamer},
  year      = {2013},
  publisher = {Bloomsbury Academic},
  address   = {London},
  edition   = {1st},
  series    = {Bloomsbury Revelations},
  isbn      = {9781780936246},
  note      = {Translated by Joel Weinsheimer and Donald G. Marshall}
}

@misc{ibrahim2025thinkinganthropomorphicparadigmbenefits,
      title={Thinking beyond the anthropomorphic paradigm benefits LLM research}, 
      author={Lujain Ibrahim and Myra Cheng},
      year={2025},
      eprint={2502.09192},
      archivePrefix={arXiv},
      primaryClass={cs.CL},
      url={https://arxiv.org/abs/2502.09192}, 
}

@misc{shanahan2023talkinglargelanguagemodels,
      title={Talking About Large Language Models}, 
      author={Murray Shanahan},
      year={2023},
      eprint={2212.03551},
      archivePrefix={arXiv},
      primaryClass={cs.CL},
      url={https://arxiv.org/abs/2212.03551}, 
}

@misc{ibrahim2025multiturnevaluationanthropomorphicbehaviours,
      title={Multi-turn Evaluation of Anthropomorphic Behaviours in Large Language Models}, 
      author={Lujain Ibrahim and Canfer Akbulut and Rasmi Elasmar and Charvi Rastogi and Minsuk Kahng and Meredith Ringel Morris and Kevin R. McKee and Verena Rieser and Murray Shanahan and Laura Weidinger},
      year={2025},
      eprint={2502.07077},
      archivePrefix={arXiv},
      primaryClass={cs.CL},
      url={https://arxiv.org/abs/2502.07077}, 
}

@article{chalmers2023conscious,
  title={Could a Large Language Model be Conscious?},
  author={Chalmers, David J.},
  journal={arXiv preprint arXiv:2303.07103},
  year={2023},
  url={https://arxiv.org/abs/2303.07103}
}

@article{goldstein2024chatgpt,
  title={Does ChatGPT Have a Mind?},
  author={Goldstein, Simon and Levinstein, Ben},
  journal={arXiv preprint arXiv:2407.11015},
  year={2024},
  url={https://arxiv.org/abs/2407.11015}
}

@article{kang2025identifying,
  title={Identifying Features that Shape Perceived Consciousness in Large Language Model-based AI},
  author={Kang, Bongsu and Kim, Jundong and Yun, Tae-Rim and Bae, Hyojin and Kim, Chang-Eop},
  journal={arXiv preprint arXiv:2502.15365},
  year={2025},
  url={https://arxiv.org/abs/2502.15365}
}

@article{vromen2024semiotic,
  title={Large Language Models as Semiotic Machines: Language Modeling vs Cognitive Modeling},
  author={Vromen, Elad},
  journal={arXiv preprint arXiv:2410.13065},
  year={2024},
  url={https://arxiv.org/abs/2410.13065}
}

@book{eco1976theory,
  author    = {Umberto Eco},
  title     = {A Theory of Semiotics},
  year      = {1976},
  publisher = {Indiana University Press},
  address   = {Bloomington},
  series    = {Advances in Semiotics}
}

@article{Salles2020,
  author    = {Arleen Salles and Kathinka Evers and Michele Farisco},
  title     = {Anthropomorphism in AI},
  journal   = {AJOB Neuroscience},
  volume    = {11},
  number    = {2},
  pages     = {88--95},
  year      = {2020},
  doi       = {10.1080/21507740.2020.1740350},
  url       = {https://doi.org/10.1080/21507740.2020.1740350}
}

@book{marcus2022rebooting,
	author = {Gary Marcus and Ernest Davis},
	editor = {},
	publisher = {Vintage},
	title = {Rebooting Ai: Building Artificial Intelligence We Can Trust},
	year = {2019}
}
\end{document}